%% file: main_arixv.tex
\documentclass[10pt,twocolumn,letterpaper]{article}

\usepackage[pagenumbers]{cvpr} % To force page numbers, e.g. for an arXiv version

\input{preamble}

\definecolor{cvprblue}{rgb}{0.21,0.49,0.74}
\usepackage[pagebackref,breaklinks,colorlinks,citecolor=cvprblue]{hyperref}
\usepackage{graphicx}
\usepackage {soul}
\usepackage{changes}
\usepackage{pifont}
\usepackage{multirow}
\usepackage{array}
\usepackage{booktabs}
\def\paperID{516} % *** Enter the Paper ID here
\def\confName{CVPR}
\def\confYear{2024}

\title{DocStormer: Revitalizing Multi-Degraded Colored Document Images to Pristine PDF}

\author{
	Chaowei Liu{$^{12}$}\thanks{This work was conducted during an internship at Huawei Technologies Co., Ltd.} 
	\quad Jichun Li{$^{1\dag}$} 
	\quad Yihua Teng{$^{1}$} 
	\quad Chaoqun Wang{$^{1}$} 
	\quad Nuo Xu{$^{1}$} 
	\quad Jihao Wu{$^{1}$} 
	\quad Dandan Tu{$^{1}$} 
	\\
	\vspace{6pt}
	{$^{1}$}Huawei Technologies Ltd,~~~ {$^{2}$}National University of Singapore.
}

\begin{document}
\maketitle
\input{sec/0_abstract} 
\input{sec/1_introduction}   
\input{sec/2_related_works}     
\input{sec/3_Proposed_Method}

\input{sec/3_dataset}

\input{sec/4_experiments}

\input{sec/5_conclusion}
\clearpage
{
	\small
	\bibliographystyle{ieeenat_fullname}
	\bibliography{main}
}


\end{document}

%% file: preamble.tex
\usepackage[dvipsnames]{xcolor}

%% file: sec/0_abstract.tex
\begin{abstract}
	For capturing colored document images, e.g. posters and magazines, it is common that multiple degradations such as shadows, wrinkles, etc., are simultaneously introduced due to external factors. Restoring multi-degraded colored document images is a great challenge, yet overlooked, as most existing algorithms focus on enhancing color-ignored document images via binarization. Thus, we propose DocStormer, a novel algorithm designed to restore multi-degraded colored documents to their potential pristine PDF. The contributions are: firstly, we propose a ``Perceive-then-Restore'' paradigm with a reinforced transformer block, which more effectively encodes and utilizes the distribution of degradations. Secondly, we are the first to utilize GAN and pristine PDF magazine images to narrow the distribution gap between the enhanced results and PDF images, in pursuit of less degradation and better visual quality. Thirdly, we propose a non-parametric strategy, PFILI, which enables a smaller training scale and larger testing resolutions with acceptable detail trade-off, while saving memory and inference time. Fourthly, we are the first to propose a novel Multi-Degraded Colored Document image Enhancing dataset, named MD-CDE, for both training and evaluation. Experimental results show that the DocStormer exhibits superior performance, capable of revitalizing multi-degraded colored documents into their potential pristine digital versions, which fills the current academic gap from the perspective of method, data, and task.
\end{abstract}

%% file: sec/1_introduction.tex
\section{Introduction} %% 打印、查看
Among various scenarios, capturing colored document images from sources such as books, posters, and magazines is particularly significant. In this context, both text and patterns (color and texture) carry essential information and are integral to creating a comprehensive digital representation of the physical document. Unfortunately, several types of degradation, such as uneven illumination, wrinkles, ink bleed-through etc., can be introduced simultaneously into the captured color document images (CDIs) due to external factors, like paper material and lighting conditions. This phenomenon significantly influences subsequent downstream tasks, like Optical Character Recognition (OCR), layout analysis, etc. Thus, sophisticated restoration algorithms for multi-degraded color document images (MD-CDIs) are needed to produce a clear, enhanced version, to ideally resemble their potential pristine digital image versions.\par

\begin{figure}
	\centering
	
	% 第一行图片
	\begin{minipage}{.3\linewidth}
		\includegraphics[width=\linewidth]{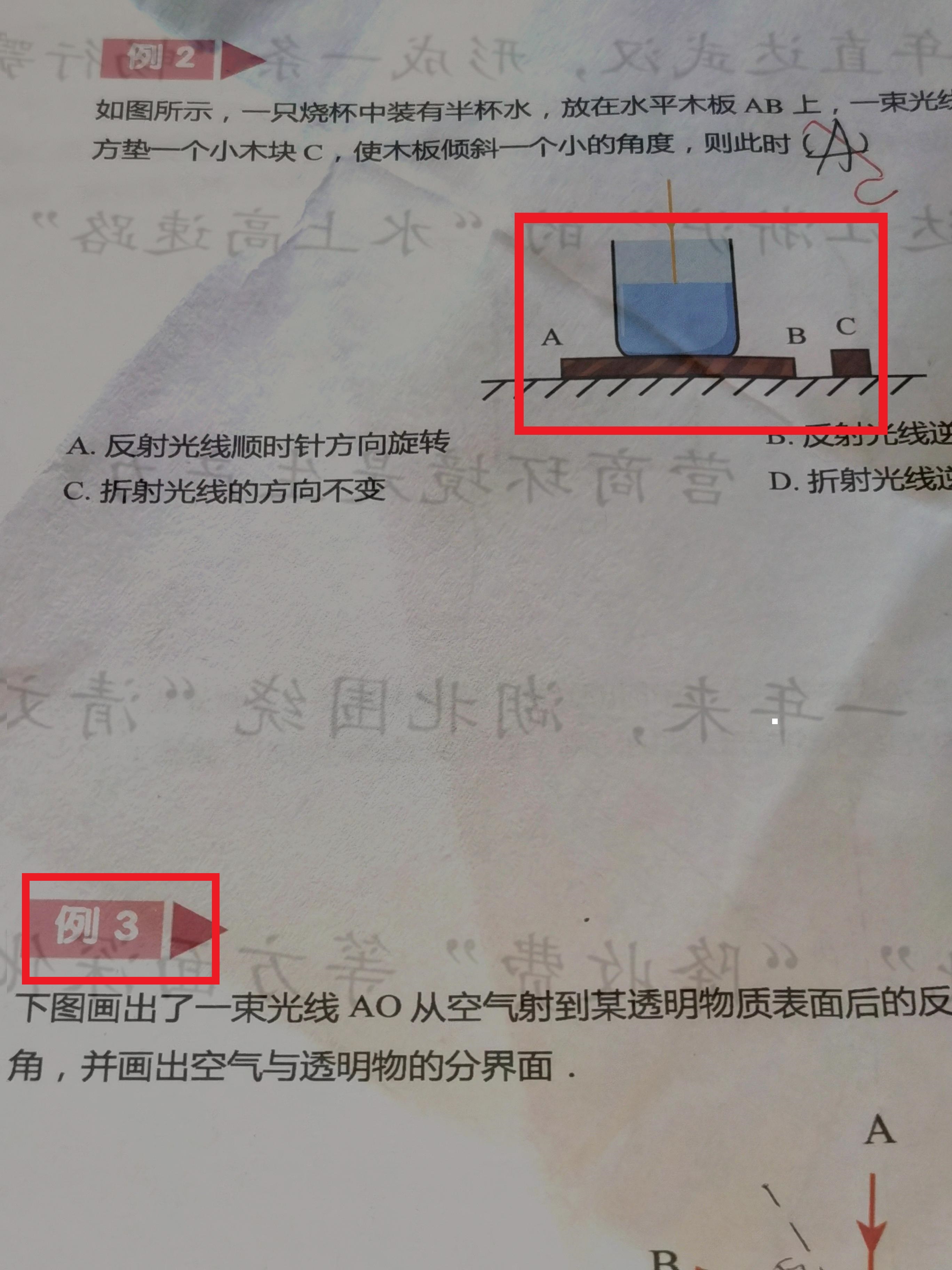}
		
	\end{minipage}%
	\begin{minipage}{.3\linewidth}
		\includegraphics[width=\linewidth]{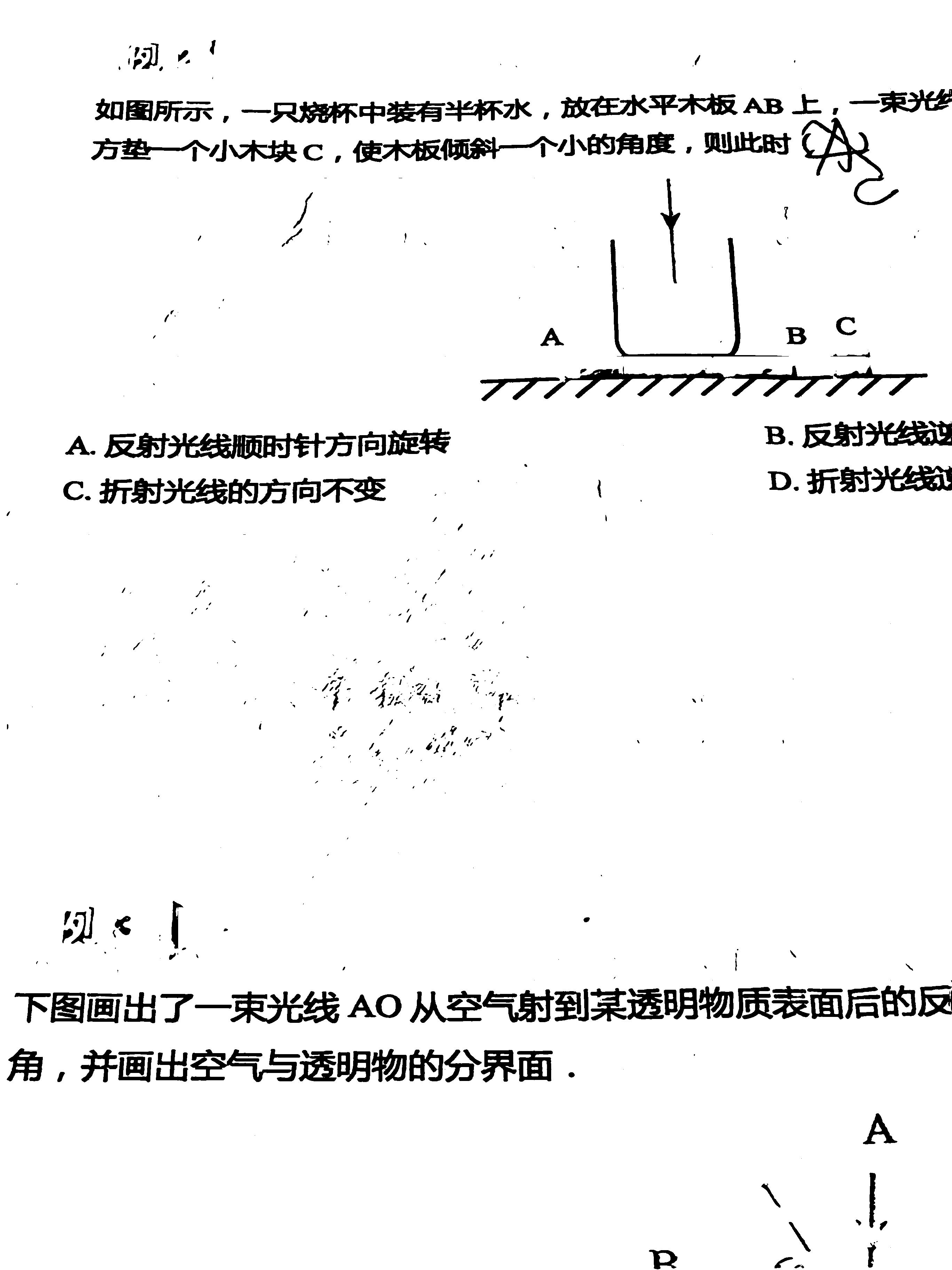}
		
	\end{minipage}%
	\begin{minipage}{.3\linewidth}
		\includegraphics[width=\linewidth]{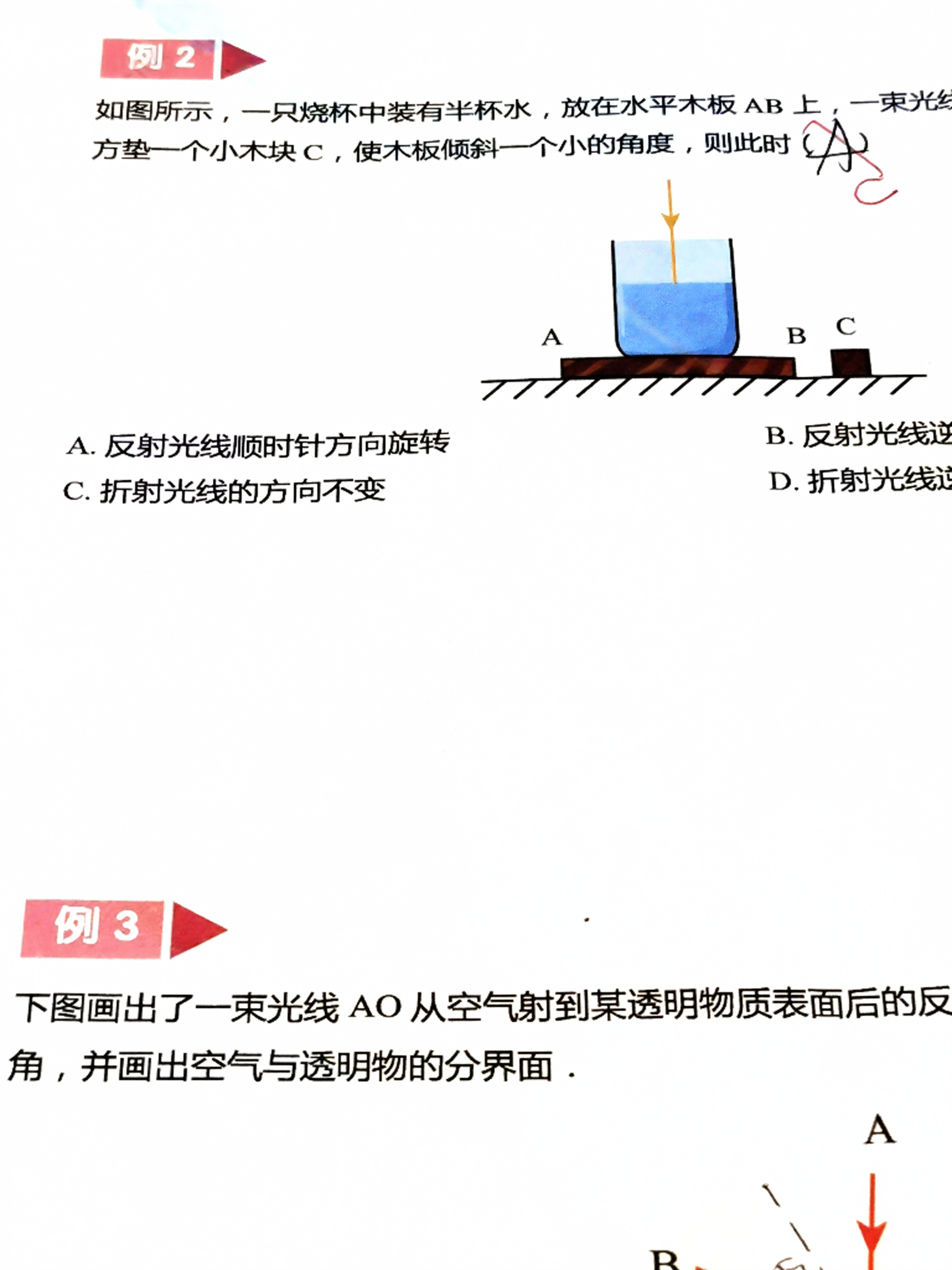}
		
	\end{minipage}
	
	% 第二行图片
	\begin{minipage}{.3\linewidth}
		\includegraphics[width=\linewidth]{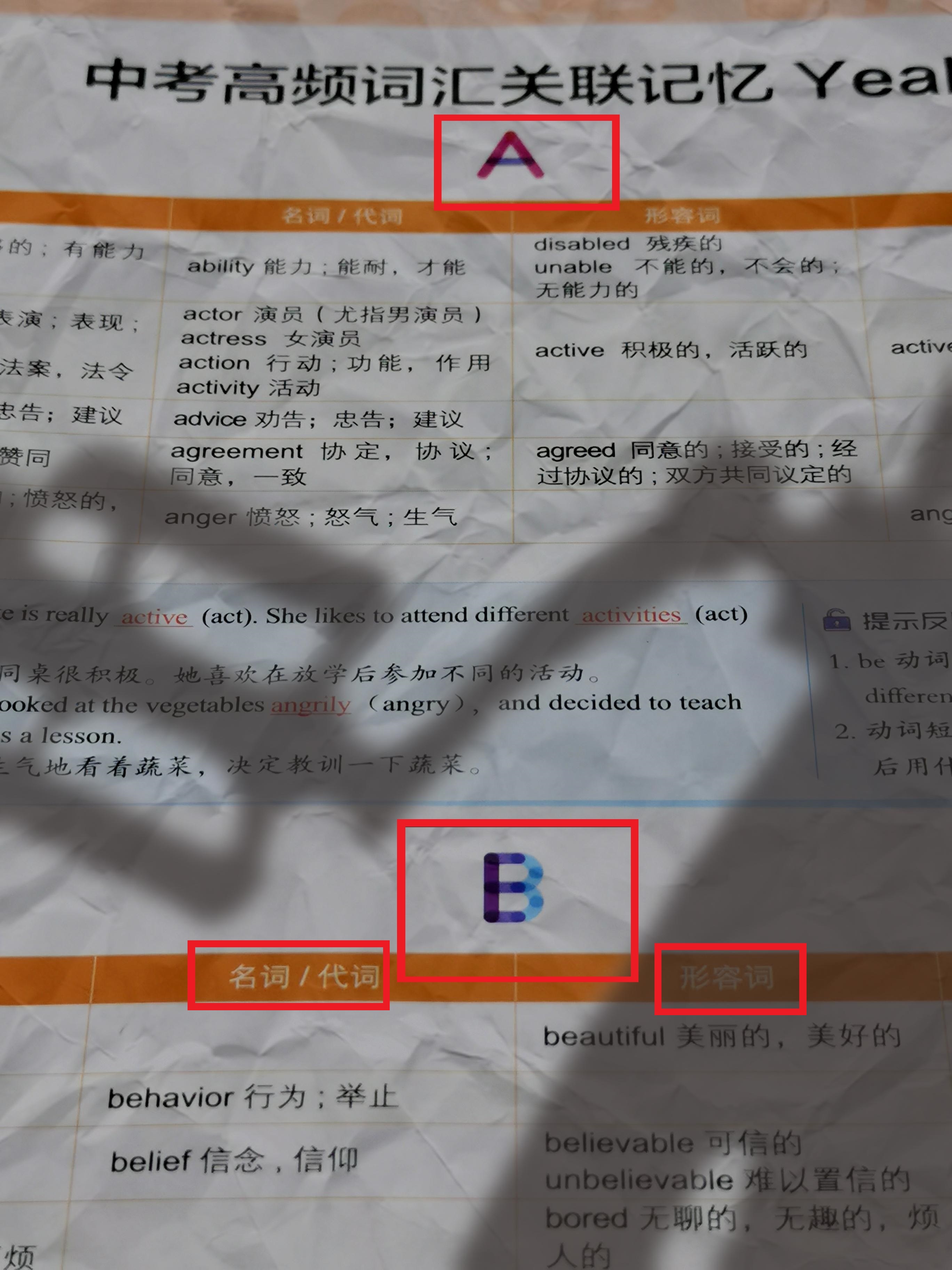}
		\subcaption{Multi-Degraded}
	\end{minipage}%
	\begin{minipage}{.3\linewidth}
		\includegraphics[width=\linewidth]{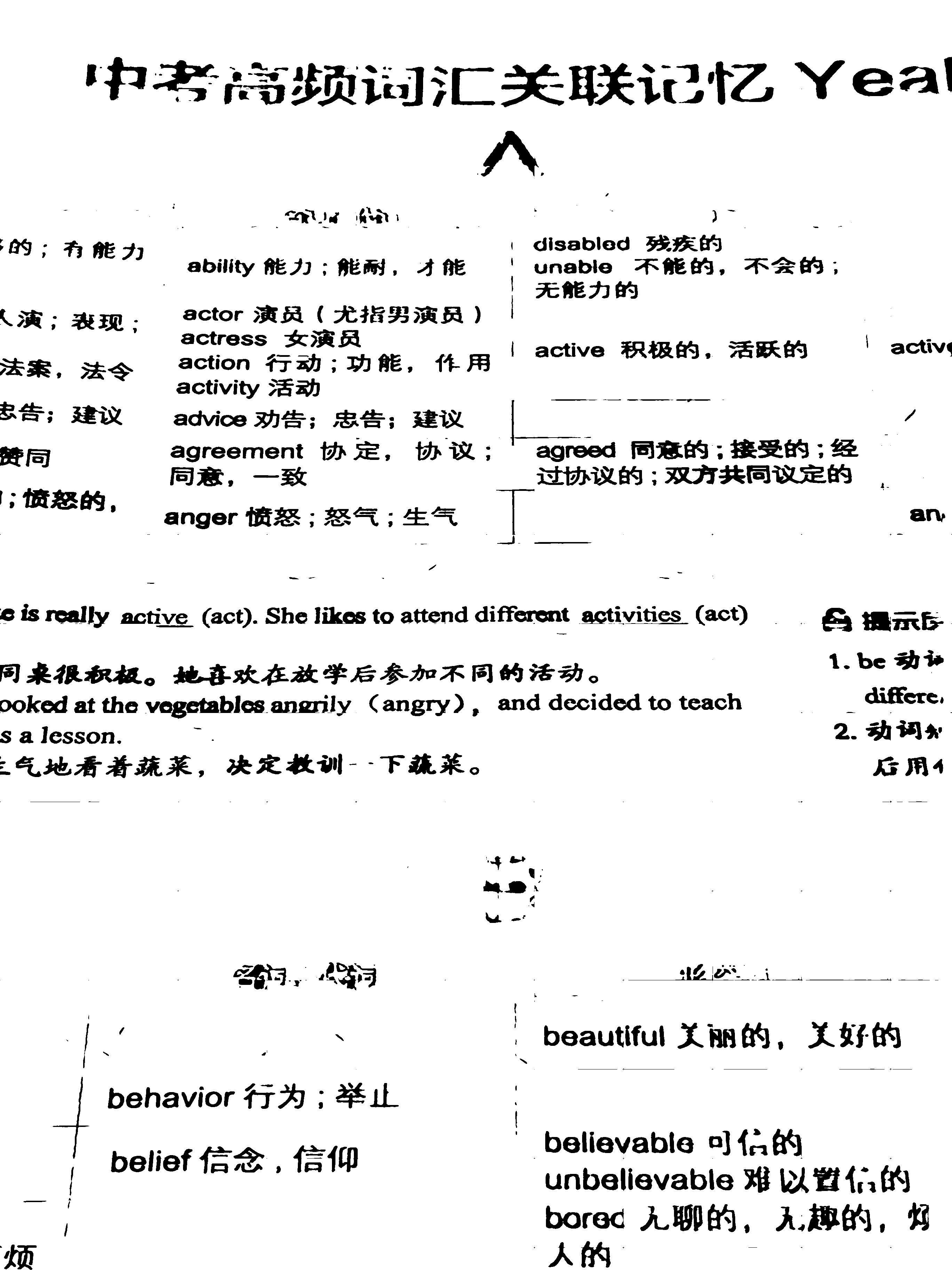}
		\subcaption{DE-GAN\cite{souibgui2020gan}}
	\end{minipage}%
	\begin{minipage}{.3\linewidth}
		\includegraphics[width=\linewidth]{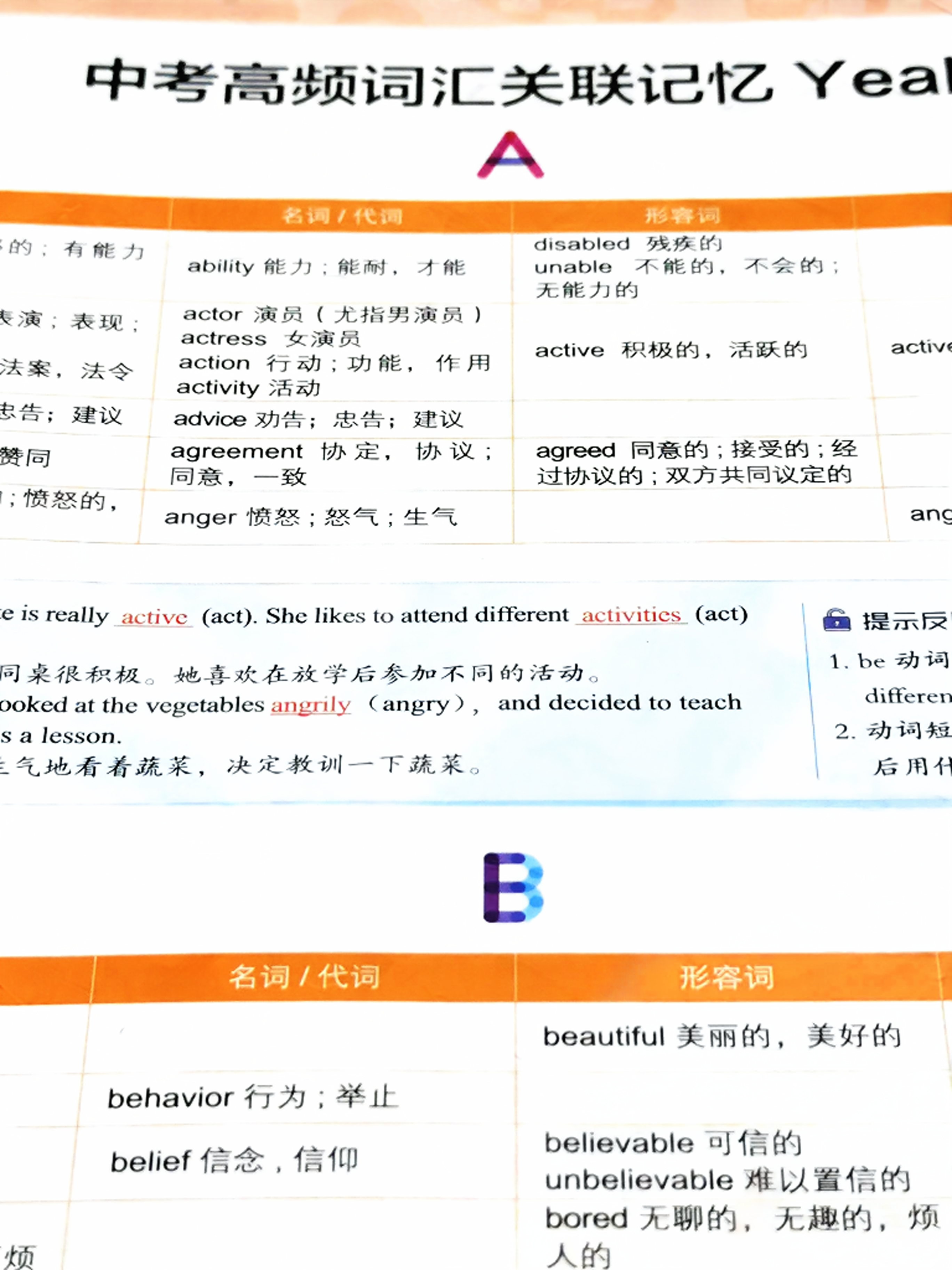}
		\subcaption{Ours}
	\end{minipage}
	
	\caption{The limitation of Multiple Degraded Colored Document Image (MD-CDI) enhancement via binarization. In the red box, the essential color pattern with semantics (liquid in the beaker) and even the text content of the paragraph title is lost during binarization, which are discarded as meaningless background or noise.}
	\label{fig:intro_demo}
\end{figure}

MD-CDIs enhancement is much more challenging than document image binarization, which goes far beyond merely increasing channels to three. They are indeed different tasks. Firstly, for binarization, the primary objective is simply to determine whether each pixel is textual foreground or not, making it a pixel-wise 0/1 classification task. In contrast, for MD-CDIE, it's not about discarding non-foreground textual pixels. Instead, it discerns and locates multiple degradations intertwined with colored textures, and it should apply different recovery strategies, depending on the type and intensity of the degradation.

Secondly, the presence of multiple degradations and their interactions with colored patterns elevate the restoration challenge, necessitating semantic completeness and consistency in the results. As shown in Fig.\ref{fig:intro_demo}, the essential color pattern and even the text information can be lost during binarization, which carries important semantic information for the readers. Thirdly, in the natural image restoration field, most of them focus on removing only one kind of degradation at a time \cite{li2022all,potlapalli2023promptir,zamir2022restormer,zamir2021multi}, which treat each degradation in isolation, failing to consider the interplay and unique characteristics of each degradation.\par

In this paper, we propose DocStormer, a transformer-based algorithm revitalizing document images to their potential pristine digital version. It can identify and eliminate various types of co-existing degradations, showing scalability towards additional degradations that may emerge in the future demand, with no more inductive bias. Moreover, it can remove various degradations while preserving original semantics, and intensifying the vibrancy of both text and color regions. The main contributions are as follows:

\begin{itemize} 
	\item We introduce a Perceive-then-Restore model, analyzing and eradicating concurrent degradations using heuristics, generalizing to possible new degradation types.
	\item We are the first to utilize GAN and pristine PDF images to narrow the distribution gap between the result and PDF, resulting in less degradation and better quality.
	\item We propose the Retinex-based strategy, PFILI, which allows for small scale training while large scale inference, It has no additional parameters and significant performance drop, while saving memory and inference time.
	\item We are the first to propose a novel MD-CDIs restoration dataset, MD-CDE, for both training and evaluation.
	\item Experimental results show that the DocStormer exhibits superior performance, and our research fills the current academic gap of MD-CDI enhancement from the perspective of method, data, and task.
\end{itemize}

%firstly, we propose a perceive-then-restore paradigm with a reinforced transformer block, which more effectively perceives and encodes the distribution of degradations.  --- 能够兼容更多的潜在的劣化类型并且不需要特别的inductive bias 设计。

%% token携带这啥信息，融合注意有利于什么的提起 -- 分别对应 ASTM STM
%% bentch mark 和dataset，前者可以不可以仅仅表示测试集

%%%%%%%% 一定要加上
%% 没有充分挖掘每一种劣化的分布特点。  distribution characteristics of each degradation.
%% 既有针对性，又有扩展性。即针对不同劣化做不同的感知，同时又没有inductive bias，还能应对潜在的新型劣化。
%  analysing and eliminating for each of the co-existing degradations respectively,

%%%%%%% 好概括
%\item We propose the Retinex-based strategy, PFILI, which allows for training on images with smaller resolution scales, yet is adaptive to testing on larger scales without additional parameters and significant performance loss, while saving memory and inference time.

%% file: sec/2_related_works.tex
\section{Related works}
\subsection{Document Image Enhancement}
\label{related_work:DocumentImageEnhancement}
Document image enhancement (DIE) has long been active in the document analysis community. The early methods relied on thresholding, identifying either a single global or multiple local threshold values to categorize document image pixels into foreground (black) or background (white)\cite{otsu1979threshold,sauvola2000adaptive}. The degradation by then is easy and always discarded together with the background. Xiong \emph{et al.} \cite{xiong2018degraded} propose support vector machine (SVM) based method, while Hedjam \emph{et al.} \cite{hedjam2014constrained} propose an energy function optimizing method tracking text pixels.\par
More recently, CNN-based methods began to take place\cite{afzal2015document,lore2017llnet,calvo2019selectional,akbari2020binarization,tensmeyer2017document}. In the method \cite{souibgui2020gan}, a conditional GAN approach is proposed for different enhancement tasks achieving good results in document binarization, deblurring, etc. \cite{souibgui2022docentr} presents a new transformer-based auto-encoder to enhance both machine-printed and handwritten document images. Doc-diff \cite{yang2023docdiff} is the first diffusion-based enhancing algorithm where a diffusion subnet is used in generating high-frequency information. \par
Currently, document images are no longer confined to monochrome historical documents but more frequently include colored document content like posters, magazines, etc. For MD-CDI enhancement (MD-CDIE), color information often carries important semantics for readers. However, current DIE methods fail to maintain such essential information, or even text characters, due to the complexity of color and degradation distribution, as shown in Fig.\ref{fig:intro_demo}. As for optimization targets, traditional DIE uses human-annotated text, while MD-CDIE uses human-edited images using Photoshop, referencing colored digital PDF publications.

\subsection{Natural Image Restoration}
Natural image restoration aims to recover clear and accurate content from degraded natural scene images. Recently, CNN-based image restoration \cite{vaswani2017attention,dudhane2022burst,zamir2022learning,zamir2021multi,anwar2020densely,zhang2020residual,wang2022uformer,zhang2021plug,liang2021swinir} excel conventional restoration approaches \cite{hek2011singleimagehazeremovalusingdarkchannelprior,kopf2008deep,timofte2013anchored}.\par
Recently, \cite{zamir2021multi} progressively learns restoration by utilizing contextualized features, high-resolution branches, and adaptive design. \cite{wang2022uformer} introduces a locally-enhanced window Transformer block and a learnable multi-scale modulator for better results. \cite{li2022all} is an all-in-one solution for single degradation of unknown type, without heuristics. Restormer \cite{zamir2022restormer} advances beyond the aforementioned methods by incorporating proposed building blocks utilizing depth-wise conv so that it can capture long-range dependencies on channels' perspective. \par
It can be concluded that all of the current natural image restoration methods are designed for single degradation, ignoring the interplay between those degradations, and the unique characteristics of each type of degradations, inevitably leading to issues in complex scenarios with multiple co-existing degradations.

% 多数都是单劣化，没有考虑单种劣化的分布特点以及多种劣化与内容之间之间互相影响，必然在复杂的分布场景中出现问题。

%% mstnet mst rest

%%%%%%%%%%%%%
%% for fair comparison, all the other sotas are changed to 3 output channels.

%% 对于阴影这种劣化，阴影是全真实的、字透部分是真实的，和褶皱是合成的

%% file: sec/3_Proposed_Method.tex
\begin{figure*}[t]
	\centering
	\includegraphics[width=0.95\linewidth]{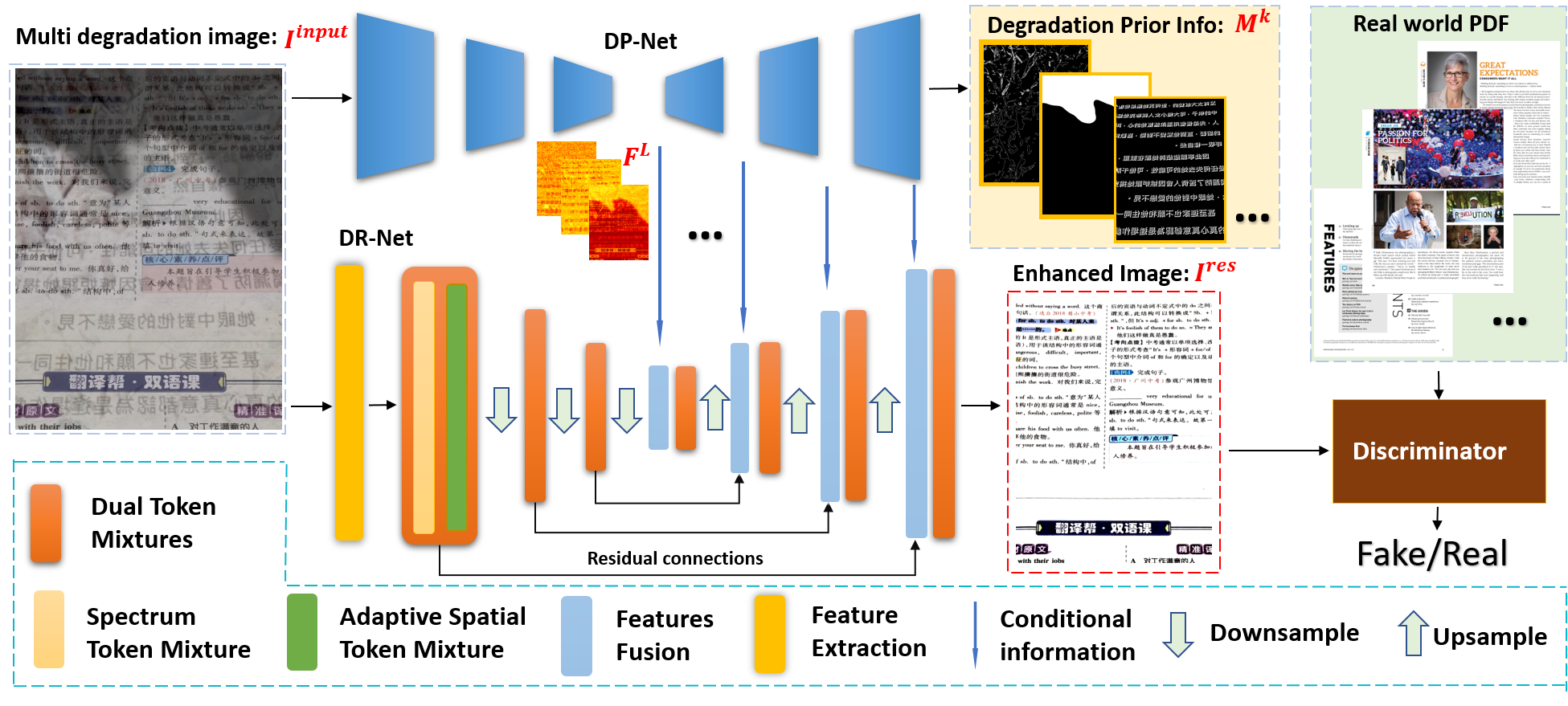}
	\caption{Overview of DocStormer with ``perceive-then-remove'' paradigm. DP-Net extracts perceptive representations for multiple-coexistent degradations in MD-CDI, which then guides DR-Net for effective degradation removal and overall enhancement. Moreover, the enhanced image is refined using Discriminator, to be more like real PDF.}

	\label{fig:overview}
\end{figure*}

\section{Proposed Method}
\label{sec:proposedMethod}
\subsection{Overview}
Our primary objective is to mitigate multiple concurrent degradations in colored document images, enhancing them to narrow the gap with authentic PDF-colored images. Figure \ref{fig:overview} depicts the architecture's overview.\par
We introduce a ``Perceive-then-Restore'' architecture: in the ``Perceive'' stage, we employ a perception subnet to extract prior degradation information. In the ``Restore'' stage, a restoration subnet is used to eliminate the degradation, leveraging the previously extracted degradation information. We employ adversarial learning to ensure the model's output closely resembles the quality and characteristics of authentic PDF images. We also introduce PFILI, a Parameter-Free Inference Method for Larger Images to facilitate scale training and large-scale inference and ensure rapid processing with minimal information loss.

Later, we will introduce four key contributions of DocStormer: (a) Perceive-then-Restore architecture, (b) Dual Transformer Block, (c) PDF-like distribution fitting via WGAN, (e)Parameter-Free Inference for Larger Images.
\subsection{Perceive-then-Restore Architecture}
\begin{figure}
	\centering
	\includegraphics[width=1\linewidth]{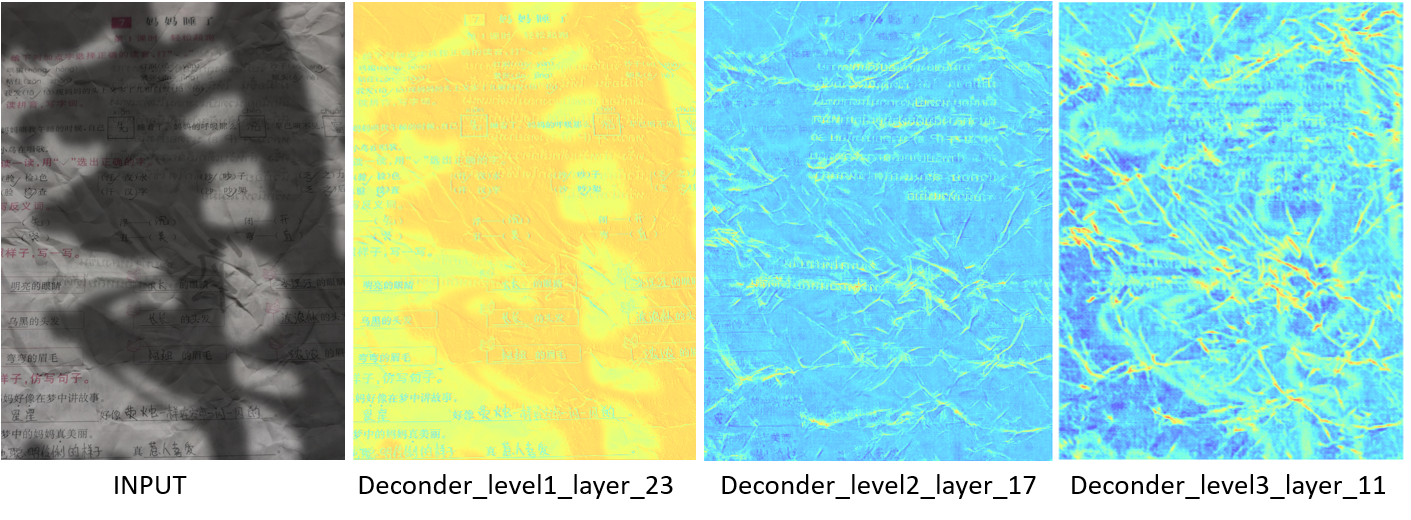}
	\caption{Visualization of Multi-Deg perceptive representation produced by the DP-Net. The term \texttt{Decoder\_levelx\_layery} denotes the $y^{th}$ layer output feature from level $x$ in decoder of the DP-Net. It is evident that this learned representation successfully encodes the intensity and position of each degradation, like shadows ($2^{nd}$ image), ink bleed ($3^{rd}$ image), and wrinkles ($3^{th}$ image).}
	\label{fig:conditioninformation}
\end{figure}
Researchers \cite{potlapalli2023promptir,guo2023shadowdiffusion} have already demonstrated that learning heuristic information depicting degradations can significantly contributes to the restoration task. Hence, we propose the ``Perceive-then-Restore'' paradigm, which first identifies the degradation type and distribution (intensity and position), and then addresses them, offering a more effective solution for restoring MD-CDIs.

In the ``Perceive'' phase, we introduce a Degradation Perception subnet (DP-Net) to discern the intensity and positions of degradation. DP-Net, a U-Net-based network, incorporates transformer attention at each layer \cite{zamir2022restormer}. We utilize features from the intermediate layers of the DP-Net's Decoder as the encoded heuristic perceptive representation.

Given an input image with multiple co-existent degradations \(I^{input}\), there are corresponding single-channel prior features for \(T\) types of degradation, denoted as \(M^K\), where \(K \in \{1, 2, \ldots, T\}\). These priors depict the position and intensity distribution of each degradation. In this context, DP-Net takes image \(I^{input} \) as input and outputs a \(k\)-channel feature, aiming to approximate \(M^K\). The encoder's intermediate features are named Multi-Deg Perceptive Representation (denoted as \(F^L\) in Fig \ref{fig:conditioninformation}), which serves as the encoded feature, adaptively learned for multiple degradations. Here, \(L\) represents the downscaling factor with \(L \in \{1,2,4,8\}\).

In ``Restore'' phase, we introduce a Multi-Degradation Restoration and Enhancement subnet (DR-Net), which adopts a U-Net-like structure and incorporates our proposed Dual Transformer Block, as detailed in Section \ref{sec:Dual Transformer Block}.

The network takes an image \( I^{input} \) as its input. Within its decoder section, perceptive representation \( F^L \) are integrated into the network via a ``Features Fusion'' module, where outputs from the corresponding encoder layer, the previous decoder layer, and \( F^L \) are concatenated. The network's final output is the enhanced image \( I^{Res} \), which is free from degradations.

Although the segmentation networks are powerful \cite{kirillov2023segment}, it is inadequate to be adopted as DP-Net since it only captures semantic position, ignoring essential degradation intensity distribution (e.g. shadow gradient). Thus we use our own DP-Net in the 'Perceive' phase, to extract perceptive representation. Moreover, DocStormer outperforms the direct prior-input concatenation method (detailed in the supplementary) for its better analysis and exploitation of the degradation priors.

In conclusion, DP-Net adeptly identifies the degradations' intrinsic features and spatial relationships, enabling the Dual Transformer Block-based DR-Net to remove degradation effectively.

\begin{figure}[t]
	\centering
	\includegraphics[width=0.85\linewidth]{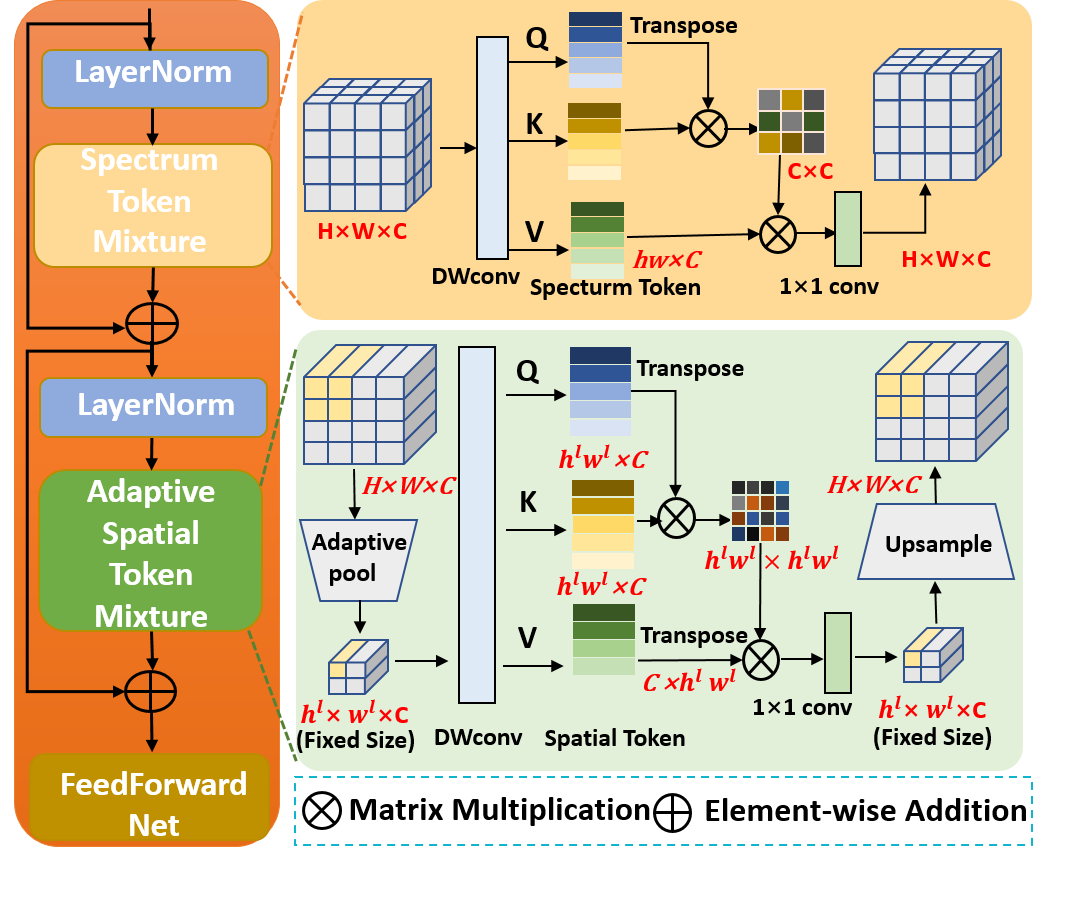}
	\caption{The overview of Dual Transformer Block }
	\label{fig:dualtransformer}
\end{figure}
\begin{figure*}[t]
	\centering
	\includegraphics[width=1\linewidth]{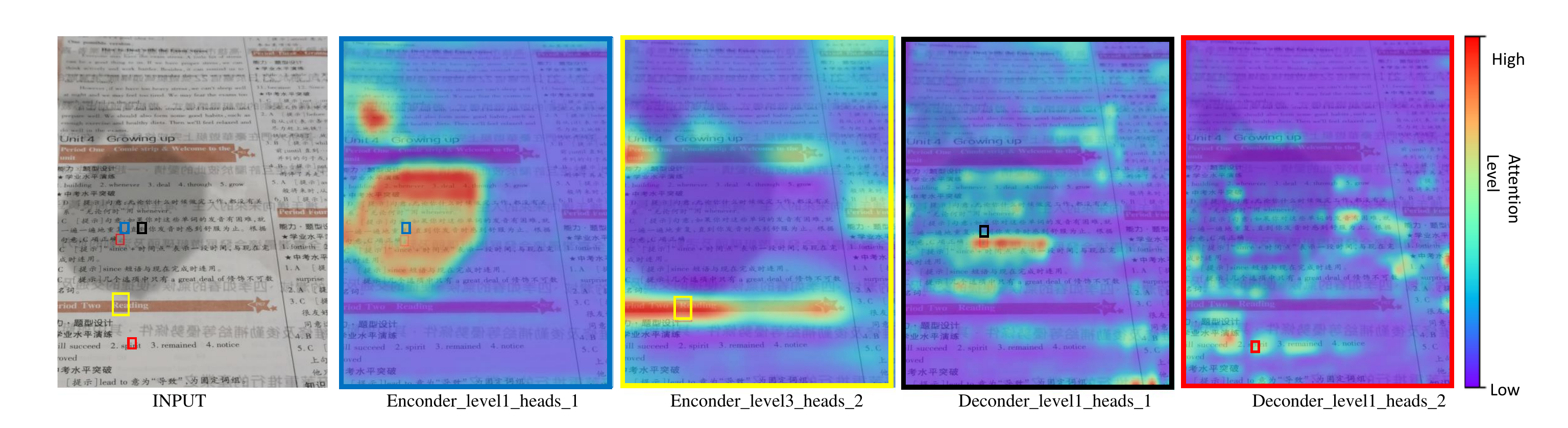}
	\caption{Attention activation in the ASTM. Each heatmap image illustrates the extent of attention that the specified bounding box region receives from the global context. These boxes represent the original image regions corresponding to the feature map pixels at different scales. We can see in global context that the algorithm pays more attentions
	to regions informative for the specified box to get reference. The term \(Encoder\_levelx\_heads\_y\) signifies the attention of the y-th head from the x-th level  encoder of the ASTM.}
	\label{fig:spatialattention}
\end{figure*}

\subsection{Dual Transformer Block}

\label{sec:Dual Transformer Block}
For MD-CDI restoration, the attention between different channel aids the model in perceiving various spectrums of different degradations. Meanwhile, spatial attention enables essential long-range dependencies, such as color variations between shadowed and illuminated areas.
However, such attention \cite{wang2018non} has a huge computation and memory cost.

To simultaneously capture attention at the inter-spectrum and spatial levels without quadratic computational growth, we introduce Dual Transformer Block (Fig.\ref{fig:dualtransformer}), which incorporates two key attention fusion modules: \textit{Spectrum Token Mixture} (STM), fusing attention across feature spectrums; \textit{Adaptive Spatial Token Mixture} (ASTM), focusing on global context attention by fixing the feature size through adaptive pooling, thus leading to linear growth in computational complexity. Fig \ref{fig:spatialattention} shows the learned ASTM attention.

For ASTM, the input features \( X_{\text{in}}^{s} \in \mathbb{R}^{ H \times W \times C } \) undergo adaptive average pooling, resulting in a downscaled fixed feature map \( X_{\text{in}}^{s\downarrow} \in \mathbb{R}^{ h^L \times w^L \times C } \).

Subsequently, we use depthwise separable convolutions to obtain \( Q_s^{\downarrow} \in \mathbb{R}^{C \times h^L w^L } \), \( K_s^{\downarrow} \in \mathbb{R}^{ h^L w^L \times C } \), and \( V_s^{\downarrow} \in \mathbb{R}^{ C \times h^L w^L } \). The dot-product of \( Q_s \) and \( K_s \) yields a fixed-size attention map \( A_s \in \mathbb{R}^{ h^L w^L \times h^L w^L } \), reducing computational complexity , as \( h^L \ll H \) and \( w^L \ll W \).

The ASTM process can be formally defined as follows:
\begin{equation}
	\begin{aligned}
		&X_{\text{out}}^{s} = \text{LN}(U^{\uparrow}(X_{\text{out}}^{s\downarrow})) + X_{\text{in}}^{s}, \, \text{where}\\
		&X_{\text{out}}^{s\downarrow} = W \left( \text{Attention} \left( Q_s^{\downarrow}, K_s^{\downarrow}, V_s^{\downarrow} \right) \right) \, \text{and}\\
		&\text{Attention}(Q,K,V) = \text{Softmax} \left( \frac{QK^T}{\sqrt{d_{k}}} \right) V.
	\end{aligned}
\end{equation}

\begin{figure}[t]
	\centering
	\includegraphics[width=1\linewidth]{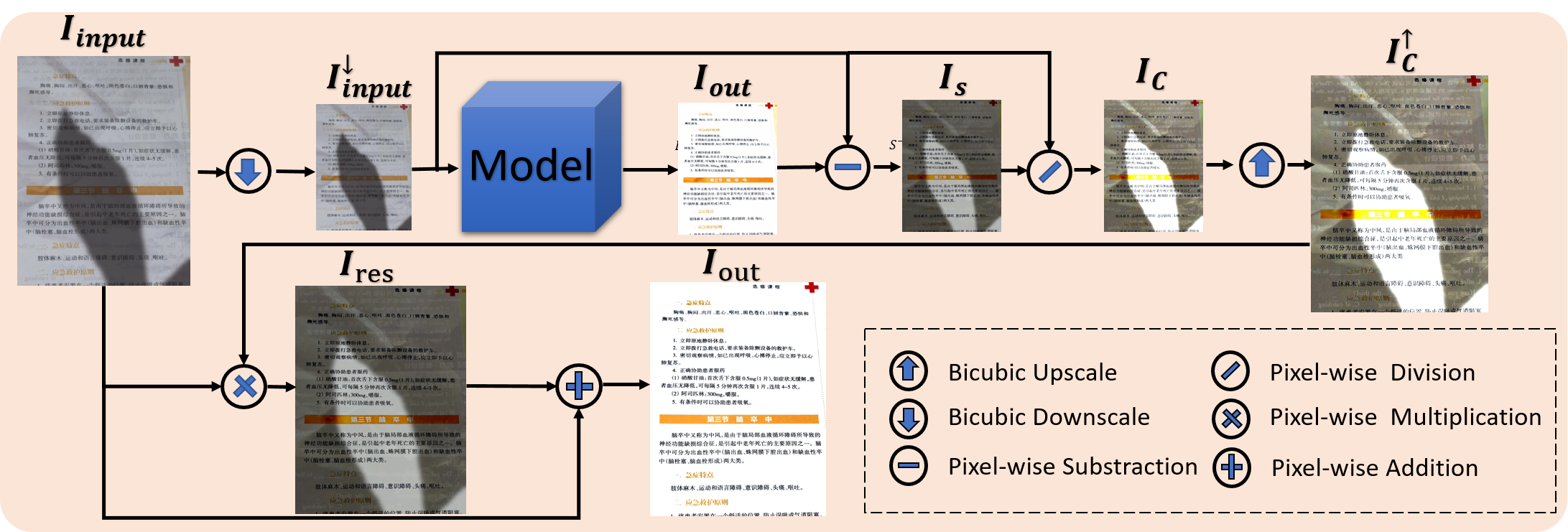}
	\caption{The overivew of PFILI, a parameter-free  inference method for larger images inference with regular image training.}
	\label{fig:rentenix}
\end{figure}

In the above equations, \(\text{LN}\) denotes Layer Normalization \cite{ba2016layer}, \(U^{\uparrow}\) is upsampling, \(X_{\text{out}}^{s}\) represents ASTM's output feature map. Here, \( d_{k} \) is a learnable scaling parameter. W is the 1×1 point-wise convolution.

For STM, the process can be mathematically defined as:
\begin{equation}
	X_{\text{out}}^{c} = \text{LN}(W(\text{Attention}(Q_c,K_c,V_c))) + X_{\text{in}}^{c}
\end{equation}

Given \( X_{\text{in}}^{c} \in \mathbb{R}^{ H \times W \times C } \), we use depthwise separable convolution to extract \( Q_c \in \mathbb{R}^{ C \times HW } \), \( K_c \in \mathbb{R}^{ HW \times C } \), and \( V_c \in \mathbb{R}^{ HW \times C } \). The dot product of \( Q_c \) and \( K_c \) yields \( A_c \in \mathbb{R}^{ C \times C } \), capturing inter-spectrum relationships.

For the Feed-Forward Network component, we adopt the Gated-Dconv Feed-Forward Network \cite{zamir2022restormer}.

\subsection{PDF-like distribution fitting via WGAN}
\label{subsec:Generator and Discriminator for PDFs}
Our goal is to produce enhanced images that closely align with human preferences, such as improved color saturation and enhanced textual clarity. Therefore, we mainly focus on high-quality PDF digital images, like those in National Geographic, produced by professional photographers. These well-composed, color-graded, and aesthetically pleasing digital PDF images, are ideal targets for our enhancements. To get similar distribution of the enhanced image to that of the digital PDF image, we use a generative adversarial training strategy based on the Wasserstein Generative Adversarial Network (WGAN) \cite{arjovsky2017wasserstein}.

Specifically, we use a discriminator to narrow the distribution gap between the enhanced images from the generator (DR-Net) and clear image patches cropped from colored PDF e-books. 
First, we collected a large number of colored PDF publications (e.g., \textit{Professional Photographer, National Geographic}) and cropped them into 100k \(256 \times 256\) image patches, each denoted as \(I^{pdf}\). Then, WGAN-GP is used to optimize the discriminator: 
\begin{equation}
	\begin{aligned}
		&\max_{D} \mathcal{L}_{WGAN-GP} = \max_{D} \left( \mathcal{L}_{WGAN} + \lambda \mathcal{L}_{GP} \right), \\
		&\mathcal{L}_{WGAN} = \mathbb{E}_{\text{crop} \sim P_{\text{crop}}} \left[ D(I^{\text{pdf}}) \right] - \mathbb{E}_I \left[ D(I^{\text{res}}) \right],
	\end{aligned}
\end{equation}
where \(\mathcal{L}_{GP}\) represents the gradient penalty term. 

Subsequently, we optimized the generator with:

\begin{equation}
	\min_{G} \left( -\mathbb{E}_I \left[ D(I^{res}) \right] \right).
\end{equation}

This optimization enhances and restores MD-CDIs to more closely resemble high-quality colored PDF e-books.

\subsection{Parameter-Free Inference for Large Images}
It is common for the model trained on regular-sized images to be required, to infer larger ones. In that case, there are two commonly used non-parametric methods on hand. The first method involves downscaling the image for model input, then upscaling the output to the original size, denoted as $bicubic$. However, this can result in the loss of high-frequency details. The second method, denoted as $crop-patches$, crops the image into smaller patches for processing through the model, subsequently piecing them together to reconstruct the entire image. While effective, this approach can be time-consuming and may yield visible block effects at patch boundaries.

\begin{table*}[t]\footnotesize
	\renewcommand\arraystretch{0.9}
	\caption{Comparison on the regular resolution (1024$\times$768) test set of MD-CDE, encompassing both synthetic and real-world datasets. }
	\centering
	\begin{tabular}{lccccccccc}
		\toprule
		\multirow{2}{*}{\centering Methods} & \multicolumn{4}{c}{Synthetic Test Set (1,300 images)} & \multicolumn{4}{c}{Real World Test Set (100 images)} & \multirow{2}{*}{\centering Params $\downarrow$} \\
		\cmidrule(lr){2-5} \cmidrule(lr){6-9}
		& PSNR(dB)$\uparrow$ & SSIM$\uparrow$ & LPIPS$\downarrow$\cite{zhang2018unreasonable}  & DISTS$\downarrow$\cite{ding2020image}& PSNR(dB)$\uparrow$ & SSIM$\uparrow$ & LPIPS$\downarrow$ & DISTS$\downarrow$ \\
		\midrule
		DE-GAN\cite{souibgui2020gan} (TPAMI20) & 22.62 & 0.9168 & 0.0773 & 0.0874 & 15.28 & 0.7330 & 0.3021 & 0.2516 & 31.04M \\
		DocEnTr\cite{souibgui2022docentr} (ICPR22) & 25.92 & 0.9407 & 0.0380 & 0.0629 & 18.31 & 0.8272 & 0.1205 & 0.1589 & 66.74M \\
		DocDiff\cite{yang2023docdiff} (ACM MM23) & 25.93 & 0.9454 & 0.0340 & 0.0738 & 19.59 & 0.8458 & 0.0988 & 0.1501 & \textbf{8.20M} \\
		Restormer\cite{zamir2022restormer} (CVPR22) & 30.21 & 0.9646 & 0.0175 & 0.0362 & 24.70 & 0.9349 & 0.0484 & 0.0659 & 26.10M \\
		\midrule
		DocStormer (Ours) & \textbf{31.40} & \underline{0.9675} & \textbf{0.0125} & \underline{0.0343}  & \underline{25.90} & \underline{0.9457} & \underline{0.0388} & \underline{0.0546} & \underline{9.50M} \\
		DocStormer\_GAN (Ours) & \underline{31.07} & \textbf{0.9676} & \underline{0.0136} & \textbf{0.0311} & \textbf{26.77} & \textbf{0.9552} & \textbf{0.0368} & \textbf{0.0442} & \underline{9.50M} \\
		\bottomrule
	\end{tabular}
	\label{lab:Comparison on low resolution}
\end{table*}

In this paper, we introduce PFILI, a non-parametric inference technique for ultra-high-resolution images, inspired by \cite{wang2019underexposed,zhang2020ris}. Specifically, it makes the neural network predict residuals at a regular resolution (e.g. 720p) resolution. Then, it combines this prediction with the original larger image at its original size, thereby swiftly producing results.

Formally, for a given input image \(I_{input}\), we first downscale it, feed it into the model, and predict the residual \(I_{s}\).

\begin{equation}
	I_{s} = \text{Model}(I_{input}^{\downarrow}) - I_{out}.
\end{equation}

Next, we obtain the downscaled correction factor \(I_{C}\) by dividing \(I_{s}\), by the downscaled original \( I_{input}^{\downarrow}\), as demonstrated below,

\begin{equation}
	I_{C} = I_{s} / I_{input}^{\downarrow}.
\end{equation}

We upscale the correction factor \(I_{C}\) using bicubic interpolation and multiply it by the original \(I_{input}\) to attain the original resolution residual \(I_{res}\), Then, we add the original-sized residual to  \(I_{input}\), yielding the final result \(I_{out}\).

\begin{equation}
	I_{\text{out}} = I_{\text{input}} \cdot (I_C^{\uparrow} + 1).
\end{equation}

%% file: sec/3_dataset.tex
\section{MD-CDE dataset}
\label{sec:MD-CDE}
As discussed in Section \ref{related_work:DocumentImageEnhancement}, contemporary document images, which often contain color patterns or information, commonly suffer from various degradations. These degradations can destroy crucial color and even textual semantic information. Despite this, existing datasets predominantly focus on binarization, and often overlook the complexities introduced by the interplay of the color pattern and multiple co-existent degradations. To bridge this gap and facilitate the development and evaluation of advanced restoration algorithms, we are the first to introduce MD-CDE, a novel dataset meticulously constructed to restore MD-CDIs. \par

\textbf{Dataset Construction}
In the MD-CDE dataset, we focus on three prevalent types of degradations: wrinkles, shadows, and ink bleed-through. The rationale lies in their ubiquity in real-world scenarios and their significant impact on the quality of document images. Potential new types of degradation can be added by following the steps below.\par

First, ground truth images, captured by collection personnel using cameras ($3K\times2K$) with no degradation , undergo a two-step refinement. This refinement involves both automated code and human annotators. The code primarily erases shadows/highlights, repair and amplifies details, and adjusts brightness contrast. Next, human annotators rectify artifacts via Photoshop, referring to the visual quality of colored PDFs. Ultimately, we carefully select and duplicate 1,300 images to form a high-resolution test set; concurrently, all ground truth images are resized to $1024\times 768$, constituting our regular training and test sets. \par

Secondly, for degradation synthesis, shadow and ink bleed-through are achieved using Augraphy\cite{augraphy_paper}. Wrinkle is simulated through a linear burning process and the photography of sampled wrinkled paper. All the random controlling parameters, such as degradation intensity, position etc., are carefully randomized empirically.\par

Thirdly, as mentioned in Sec.\ref{sec:proposedMethod}, degradation priors should be obtained, which indicate the position and intensity of their types, with appropriate value ranges. Theoretically, annotators can manually use Photoshop to get degradation priors that depict the intensity and position of their type. To simplify our implementation, we first utilize the Augraphy blending map for shadow and bleeding, and the original image for wrinkles. Subsequently, we apply Sobel edge detection and thresholding to these respective maps to obtain the resultant priors (see supplementary for details).\par

Lastly, we acquired an additional 100 real-world MD-CDIs, each with a resolution of $1024\times 768$, to serve as our real-world test set. These images feature authentic co-existing degradation(s). Annotators manually repaired these images to produce the corresponding ground truth.\par

\textbf{Scale and Division}
The MD-CDE partition includes a training set of 39,590 images and three test sets: 1) synthetic regular resolution with 1,300 images; 2) synthetic high-resolution with 1,300 images; 3) real-world regular resolution with 100 images. This partitioning allows algorithms to fully leverage the voluminous labeled data. For evaluation, an algorithm's learning capability is assessed using the 1,300-image regular test set, while its generalization capability in real-world scenarios is evaluated using the 100-image captured set, providing comprehensive insight into the algorithm's efficacy across various conditions.\par
Illustrations of MD-CDE are given in the supplementary.

%% file: sec/4_experiments.tex
\begin{figure*}[t]
	\centering

	\includegraphics[width=1\linewidth]{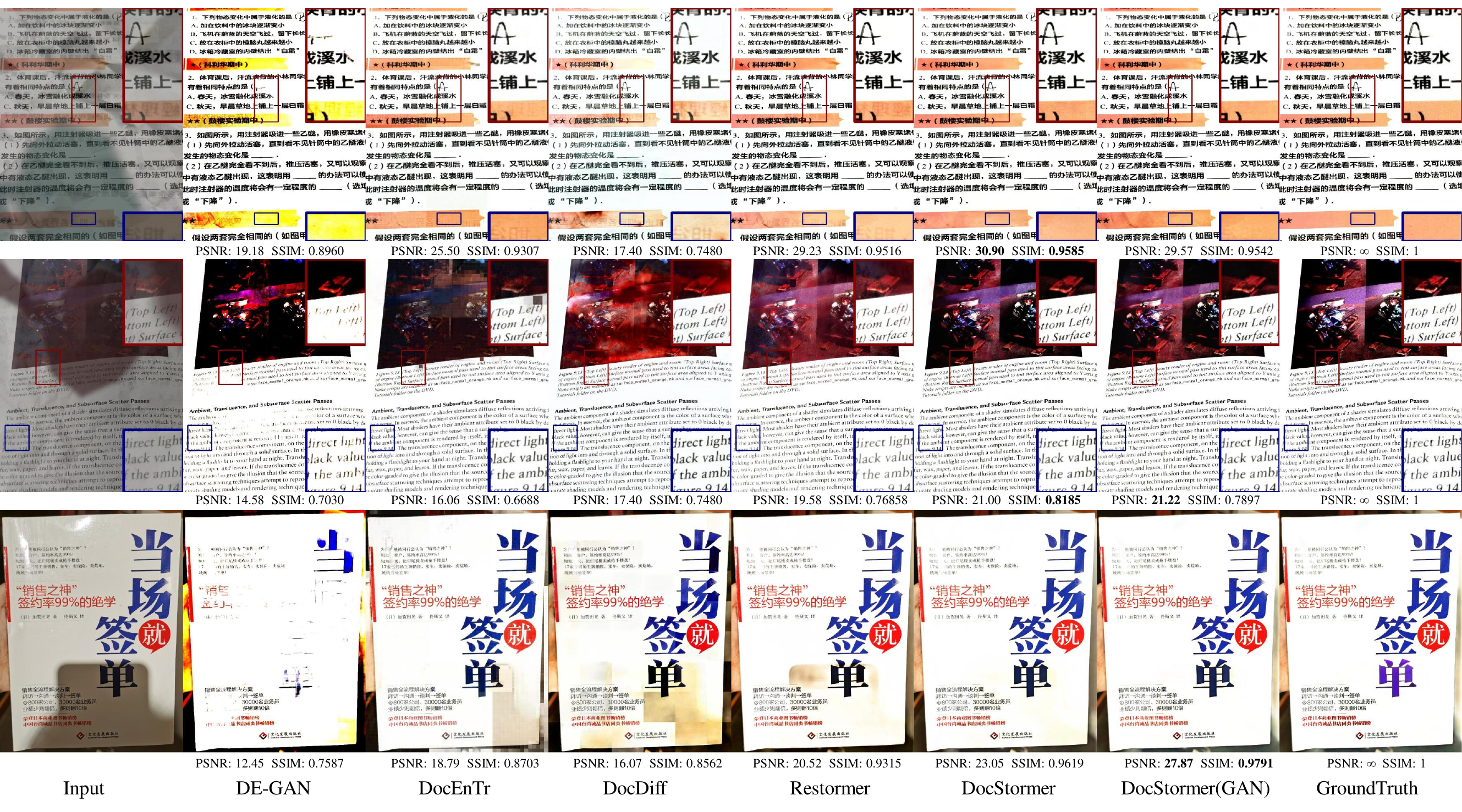}
	\caption{Visual comparison of our method with state-of-the-art approaches on the MD-CDE dataset. Our DocStormer demonstrates superior performance in both global fidelity and local detail preservation, and meanwhile DocStormer with GAN has better visual quality.}
	\label{fig:visual comparison}
\end{figure*}

\section{Experiment}
\label{sec:Experiments}

\subsection{Implementation details}
%Our model is implemented in PyTorch \cite{paszke2019pytorch} and trained on 6 Nvidia Tesla V100 GPUs. The parameters of the network are optimized with ADAM ($\beta_1 = 0.9$, $\beta_2 = 0.999$, weight decay $1e^{-4}$) \cite{kingma2014adam} algorithm for 120K iterations with the initial learning rate $3\times e^{-4}$, which is gradually reduced to $1e^{-6}$ using cosine annealing \cite{loshchilov2016sgdr}.
%For progressive learning, models are trained with 300K iterations in each stage. The patch size and batch size pairs are updated from [$192^2$,24] to [($256^2$,16), ($384^2$,8)] at iterations [100K, 150K]. For data augmentation, we use horizontal and vertical flips.
%
%Our DR-Net employs a 4-level encoder-decoder. Empirically, from level-1 to level-4, the number of Dual Transformer blocks is [2,2,2,3]. The attention heads in ASTM and STM are [1,2,2,3], and the number of channels is [36, 72, 108, 144]. The Adaptive pooling size in the ASTM is [$32^2$,$16^2$,$8^2$] from level 2 to level 4.
%
%Our DP-Net employs a 4-level encoder-decoder based on \cite{zamir2022restormer}. From level-1 to level-4, the number of Transformer blocks is [1,1,1,2]. The attention heads are [1,1,1,2], and the number of channels is [18, 36, 54, 72]. 

%% by gpt
Our model is implemented using PyTorch \cite{paszke2019pytorch} and is trained on 6 Nvidia Tesla V100 GPUs. For the optimization of network parameters, we utilize the ADAM optimizer \cite{kingma2014adam} with $\beta_1 = 0.9$, $\beta_2 = 0.999$, and a weight decay of $1e^{-4}$. Training spans 120K iterations with an initial learning rate of $3\times e^{-4}$, which is gradually decreased to $1e^{-6}$ via cosine annealing \cite{loshchilov2016sgdr}. During progressive learning, models undergo 300K iterations at each stage. We adjust patch size and batch size pairs from [$192^2$,24] to [($256^2$,16), ($384^2$,8)] at the 100K and 150K iteration marks. Data augmentation involves both horizontal and vertical flips.

Our DP-Net, built upon a 4-level encoder-decoder structure inspired by \cite{zamir2022restormer}, uses [1,1,1,2] Transformer blocks from level-1 to level-4. The attention heads are set at [1,1,1,2], with the channel numbers at [18, 36, 54, 72].

Our DR-Net features a 4-level encoder-decoder architecture. Empirically, from level-1 to level-4, we deploy [2,2,2,3] Dual Transformer blocks. The attention heads in both ASTM and STM range from [1,2,2,3], with channel numbers set at [36, 72, 108, 144]. From level 2 to level 4, the Adaptive pooling sizes in the ASTM are [$32^2$,$16^2$,$8^2$].

\textbf{Loss Functions:} We employ the Focal Loss (\( FL \)) \cite{lin2017focal} as the loss function for DP-Net. The formula for the DP-Net loss \( L_{\text{DP}} \) is defined as:
\begin{align}
	\label{eq:loss_DP}
	\mathcal{L}_{\text{DP}} &= \sum_{\text{type}} W_{\text{type}} \cdot FL_{\text{type}}, \\
	FL_{\text{type}}(p_t) &= -(1-p_t)^\gamma \log(p_t),
\end{align}
where \( p_t \) is given by:
\[
p_t = 
\begin{cases} 
	p & \text{if } y = 1, \\
	1-p & \text{otherwise.}
\end{cases}
\]
For specific degradation types, the parameter weights (\( W_{\text{type}} \)), \( p_t \), and \( \gamma \) are configured as follows: Shadow: [1, 2, 0.72], Wrinkle: [2, 2, 0.96], bleed: [3, 2, 0.94].

For training DocStormer, we load DP-Net's weight, then we jointly train DP-Net and DR-Net with $\mathcal L_{DS}$:\par
\begin{equation}
	\mathcal L_{DS}=\lambda_{1}\cdot \mathcal L_{Freq}+\mathcal \lambda_{2}\cdot L_{Color}+ \lambda_{3}\cdot\mathcal L_{1} + \lambda_{4}\cdot\mathcal L_{DP}
\end{equation} 
where L1 loss, Color Loss\cite{wang2019underexposed} $\mathcal L_{Color}$, Frequency Loss\cite{cho2021rethinking} $\mathcal L_{Freq}$ are calculated between input and ground truth. 

For training DocStormer\_GAN, firstly we also load DP-Net's weight, and then we jointly train the DocStormer and Discriminator using the below objective:
\begin{equation}
	\mathcal L_{DS\_GAN}= \lambda_{5}\cdot\mathcal L_{DS} + \lambda_{6}\cdot  L_{WGAN-GP}
\end{equation}
Empirically, $\lambda_{1}=0.1$, $\lambda_{5}=0.001$, other $\lambda$s are 1.

\begin{table*}[t]
	\centering
	\begin{minipage}[t]{.65\textwidth}
		\scriptsize
		\renewcommand\arraystretch{0.85}
		\caption{Ablation Study for DocStormer on the synthetic subset of the MD-CDE regular resolution synthetic test dataset. \ding{51} and \ding{53} indicate the inclusion and exclusion of a module, respectively. All models are trained for 300k iterations using the same experimental setup.}
		\begin{tabular}{p{0.78cm}p{1.5cm}p{0.4cm}p{0.8cm}p{1.05cm}p{0.6cm}|p{0.8cm}p{0.4cm}p{0.4cm}p{0.4cm}}
			\toprule
			\multirow{2}{*}{\centering Method} & \centering Frequency loss/Color loss& \multirow{2}{*}{\centering ASTM} & \centering With DP-Net& \centering DR-Net Pretrained& \centering With WGAN & \multirow{2}{*}{\centering PSNR(dB)$\uparrow$} & \multirow{2}{*}{\centering SSIM$\uparrow$} & \multirow{2}{*}{\centering LPIPS$\downarrow$} & \multirow{2}{*}{\centering DISTS$\downarrow$} \\                             
			\midrule
			Restormer &\centering\ding{53} & \centering\ding{53} & \centering\ding{53} & \centering\ding{53} & \centering\ding{53} &\centering30.21 & 0.9646 & 0.0175 & 0.0362 \\
			Restormer\(^\textbf{+} \)&\centering\ding{51} & \centering\ding{53} & \centering\ding{53} & \centering\ding{53} & \centering\ding{53} & \centering30.41 & 0.9658 & 0.0157 & 0.0351  \\
			Restormer\(^\textbf{++} \) &\centering\ding{51} & \centering\ding{51} & \centering\ding{53} & \centering\ding{53} & \centering\ding{53} &\centering30.89 & 0.9663 & 0.0141 & 0.0352  \\
			\midrule
			DocStormer\(^\textbf{-} \)&\centering\ding{51} & \centering\ding{51} & \centering\ding{51} & \centering\ding{53} & \centering\ding{53} &\centering31.30 & 0.9672 & 0.0131 & 0.0332  \\
			DocStormer & \centering\ding{51} & \centering\ding{51} & \centering\ding{51} & \centering\ding{51} & \centering\ding{53} &\centering\textbf{31.40} & 0.9675 & \textbf{0.0125} & 0.0343  \\
			DocStormer\_GAN& \centering\ding{51} & \centering\ding{51} & \centering\ding{51} & \centering\ding{51} & \centering\ding{51} &\centering31.07 & \textbf{0.9676} & 0.0136 & \textbf{0.0311}  \\
			\bottomrule
		\end{tabular}
		\label{label:ablation study}
	\end{minipage}%
	\hspace{0.3cm}
	\begin{minipage}[t]{.3\textwidth}
		\scriptsize
		\renewcommand\arraystretch{0.85}
		\caption{Comparison on a high-resolution test set (3k$\times$2k). TIME(s) denotes 50 images' average time on Tesla V100.}
		\begin{tabular}{p{1.9cm}p{0.8cm}p{0.5cm}p{0.5cm}}
			\toprule
			Methods & PSNR(dB)$\uparrow$ & SSIM$\uparrow$ & TIME(s)$\downarrow$ \\
			\midrule
			Restormer & & & \\
			\hspace{0.5em} $crop-patches$ & \centering24.70 & 0.9500 &8.951 \\
			\hspace{0.5em} $bicubic$ & \centering26.50 & 0.9526  &0.622 \\
			\hspace{0.5em} Ours (PRILI) & \centering27.50 & 0.9224 &1.567 \\
			\midrule
			DocStormer & & & \\
			\hspace{0.5em} $crop-patches$ & \centering25.93 & 0.9530  &10.93 \\
			\hspace{0.5em} $bicubic$ & \centering27.00 & \textbf{0.9550}  & 0.767 \\
			\hspace{0.5em} Ours (PRILI) & \centering\textbf{28.03}& 0.9242  & 1.886 \\
			\bottomrule
		\end{tabular}
		\label{label:Experiments Results on Rentenix}
	\end{minipage}
\end{table*}

\subsection{Experiments Results}

\begin{figure*}[t]
	\centering
	\includegraphics[width=1\linewidth]{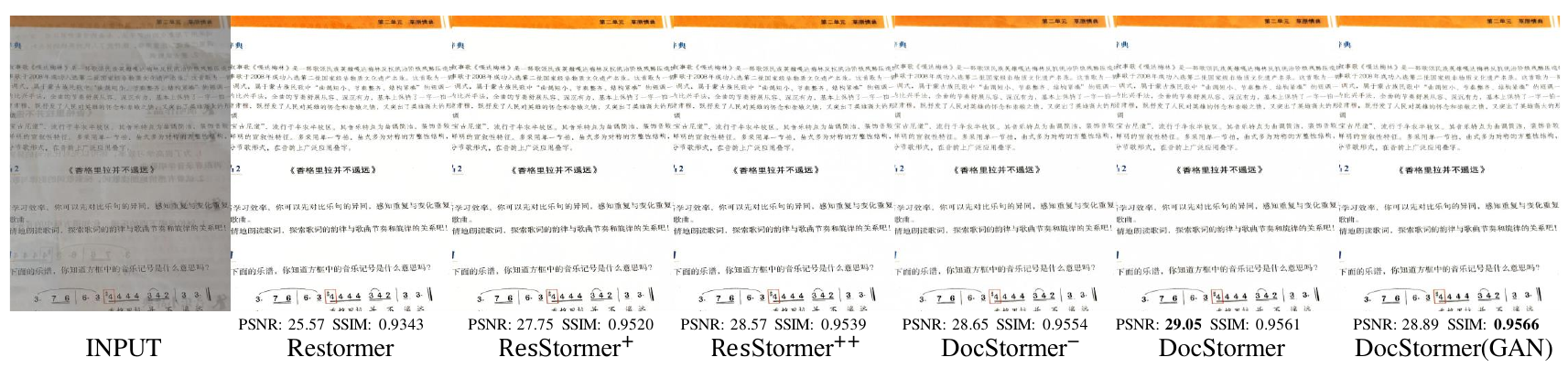}
	\caption{Visualization of ablation study. Due to the limitation, more visual results are given in supplementary material.}
	\label{fig:mainablation}
\end{figure*}

No algorithm explicitly designed for restoring multi-degraded colored document images exists definitively. The majority of extant document restoration algorithms predominantly focus on binarization. To facilitate a comprehensive and impartial comparison, we evaluate against leading algorithms in two distinct categories: natural image restoration \cite{zamir2022restormer} and document enhancement algorithms \cite{souibgui2020gan,souibgui2022docentr,yang2023docdiff}, modifying the output channels of these methods to 3.

Experiments on the MD-CDE show that DocStormer surpasses other state-of-the-art across an array of evaluation metrics while maintaining a lean model size of only 9.5M parameters. The results are elaborated in Table \ref{lab:Comparison on low resolution}. 

DocStormer is highly effective at mitigating a range of co-existent degradations, especially when faced with multiple concurrent degradations. Utilizing WGAN, our variant DocStormer\_GAN excel all the others in all the metrics in real-world dataset, drawing the output images closer to the quality of digital PDF images. Visual results are in Fig. \ref{fig:visual comparison}.

\subsection{Ablation Studies}

To assess the contributions of DocStormer's components, we conducted an ablation study as delineated in Table \ref{label:ablation study}. 

Commencing with Restormer's initial PSNR baseline of 30.21 dB, the incorporation of Color Loss/Frequency Loss yield better performance, surpassing L1 loss alone. The inclusion of the ASTM module further amplified the PSNR by 0.48 dB, underscoring its proficiency in global perception.

Implementing the Perceive-Then-Restore architecture with DP-Net boosted the PSNR. To ensure its initial enhancement capabilities during DP-Net and DR-Net joint training, we initialized DR-Net with pretrained weights, denoted as ``DR-Net Pretrained'' in Tab.2.

Utilizing WGAN improv SSIM and DISTS, despite a slight decrease in PSNR, indicating its preference for perceptual quality and semantics, over pure signal fidelity.

In summary, each component of DocStormer has a unique impact on image enhancement. 

\subsection{Experimental Results on PFILI}
PFILI, a plug-in non-parametric method, enables high-resolution inference with regular-resolution training. Tab.\ref{label:Experiments Results on Rentenix} shows the substantial improvements PFILI brings to the performance of DocStormer and \cite{zamir2022restormer}.

PFILI outperforms both $bicubic$ and $crop-patches$ in signal fidelity and visual quality, achieving this with notably lower time and memory. Furthermore, it harmonizes the network's perceptual scale between testing and training. A minor downside is a reduced SSIM score, a result of noise introduced during the integration of original scale signal.

%% file: sec/5_conclusion.tex
\section{Conclusion}
\begin{figure}[t]
	\centering
	\includegraphics[width=1\linewidth]{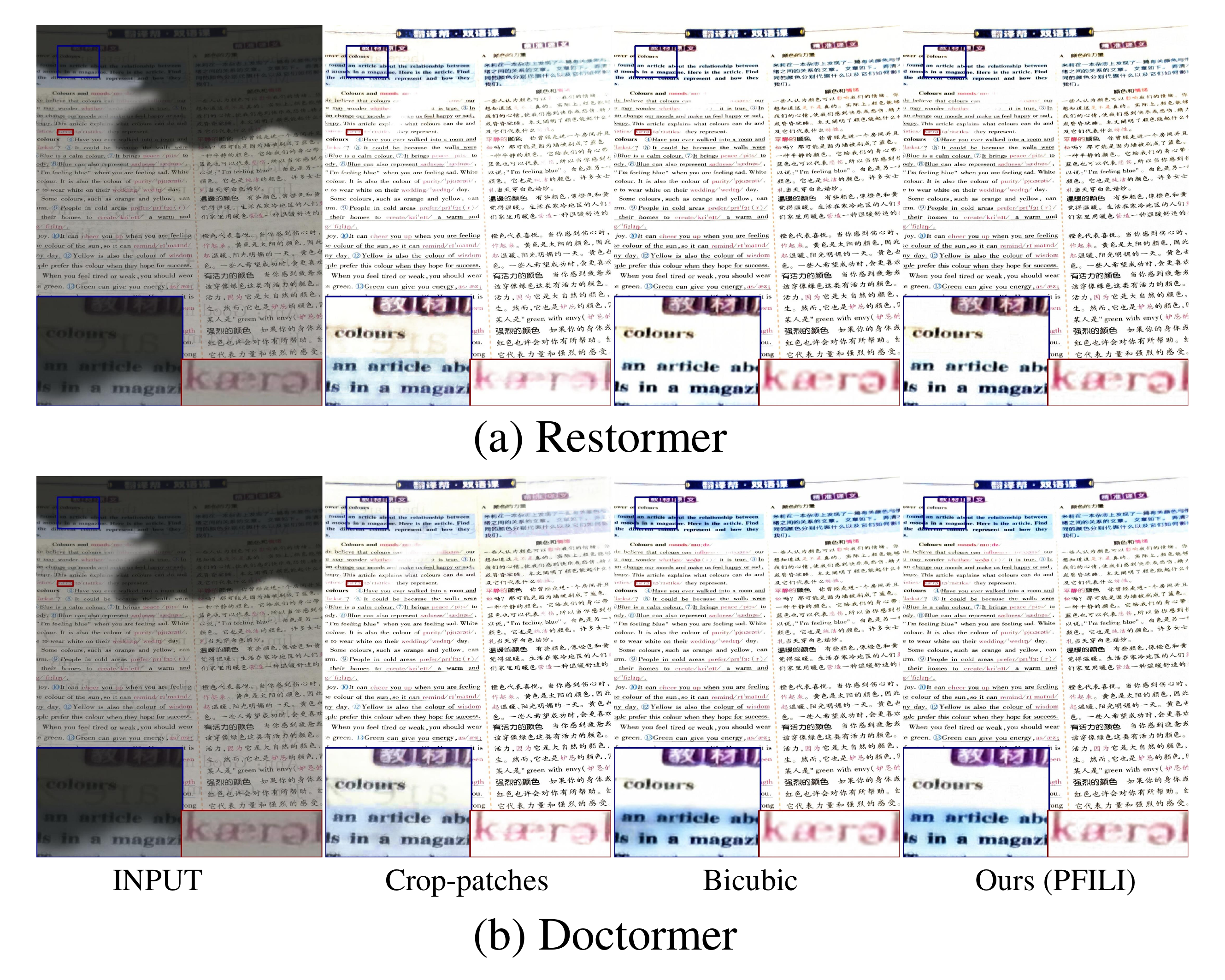}
	\caption{Visual comparison of PFILI, $bicubic$, and $crop-patches$ (applied to and assessed on DocStormer and \cite{zamir2022restormer}). PFILI ensures scale consistency and improve output quality, devoid of the block artifacts exhibited by the $crop-patches$ method. More visual results are in the supplementary material.}
	\label{fig:pfilivisuizatioin}
\end{figure}
In this paper, we propose DocStormer, an innovative algorithm that restores multi-degraded colored document images to their potential pristine digital forms, proficiently bridging existing academic and practical gaps. Firstly, through a perceive-then-restore paradigm and a reinforced transformer block, the algorithm effectively comprehends and encodes degradation distributions. Secondly, uniquely employing GANs and pristine PDF images minimizes the distribution disparity between the results and the PDFs, thereby enhancing visual quality. Thirdly, novel strategy PFILI, ensures memory and time efficiency during training and inference while compromising minimal detail. Finally, introducing the MD-CDE dataset establishes a robust foundation for future investigations and validations in MD-CDI restoration. Experimental results from synthetic and real-world data demonstrate that our algorithm excels the current state-of-the-art.